%% file: main.tex
\documentclass{article}
\usepackage{microtype}
\usepackage{graphicx}
\usepackage{booktabs}
\PassOptionsToPackage{hyphens}{url}
\usepackage{hyperref}
\makeatletter\g@addto@macro\UrlBreaks{\do\-}\mathchardef\UrlBigBreakPenalty=\@M\makeatother

\usepackage[preprint]{icml2027}

\usepackage[T1]{fontenc}
\usepackage{amsmath,amssymb}
\usepackage{multirow}
\usepackage{array}
\usepackage{enumitem}
\usepackage{pifont}
\usepackage{xspace}
\usepackage[skins]{tcolorbox}
\IfFileExists{fontawesome5.sty}{\usepackage{fontawesome5}}{\newcommand{\faGithub}{}\newcommand{\faDatabase}{}\newcommand{\faBlog}{}}

\graphicspath{{figures/}}
\newcommand{\Aset}{Clean\xspace}
\newcommand{\Bset}{Orig\xspace}
\newcommand{\AFP}{Clean+FP\xspace}
\newcommand{\BF}{Orig+Filter\xspace}
\newcommand{\SGV}{SWE-Gym-Verified\xspace}
\newcommand{\Vfour}{Verified-397\xspace}%
\newcommand{\pp}{\,pp\xspace}
\newcommand{\band}[1]{#1/8\xspace}
\newcommand{\Nset}{512\xspace}%
\newcommand{\anon}[2]{\ificmlshowauthors #2\else #1\fi}

\definecolor{takehead}{HTML}{7091C0}
\definecolor{takebg}{HTML}{F3F7FD}
\definecolor{takerule}{HTML}{2F6DB5}%
\newtcolorbox{takeawaybox}[1]{enhanced, colback=takebg, colframe=takehead, colbacktitle=takehead, coltitle=white,
  boxrule=0.6pt, arc=2pt, title={#1}, fonttitle=\bfseries,
  left=5pt, right=5pt, top=3pt, bottom=3pt, toptitle=1.5pt, bottomtitle=1.5pt, lefttitle=5pt,
  before skip=7pt plus 1pt, after skip=6pt plus 1pt}
\newcommand{\takeaway}[3]{\begin{takeawaybox}{Takeaway #1: #2}#3\end{takeawaybox}}
\newcommand{\pmstd}[1]{{\scriptsize\,$\pm$\,#1}}

\newcommand{\releaselink}[3]{\href{#1}{\textcolor{takerule}{#2\hspace{0.45em}\textbf{#3}}}}

\icmltitlerunning{What Makes Good Data for Training Coding Agents? A Systematic Study on SWE-Gym}

\begin{document}

\twocolumn[
  \icmltitle{What Makes Good Data for Training Coding Agents?\\ A Systematic Study on SWE-Gym}

  \begin{icmlauthorlist}
    \icmlauthor{Yikai Zhang}{amd}
    \icmlauthor{Mai Xu}{umich}
    \icmlauthor{Yunbei Zhang}{tulane}
    \icmlauthor{Qirui Jin}{gatech}
    \icmlauthor{Kenan Li}{ucl}
    \icmlauthor{Yuwen Jia}{ucla}
    \icmlauthor{Jinwei Yao}{amd}
    \icmlauthor{Xiangyu Zhang}{amd}
    \icmlauthor{Jielun Tan}{amd}
    \icmlauthor{Jin Pan}{amd}
    \icmlauthor{Zijian Jin}{nyu}
    \icmlauthor{Xiao Sun}{amd}
  \end{icmlauthorlist}

  \icmlaffiliation{amd}{AMD}
  \icmlaffiliation{umich}{University of Michigan}
  \icmlaffiliation{tulane}{Tulane University}
  \icmlaffiliation{gatech}{Georgia Institute of Technology}
  \icmlaffiliation{ucl}{University College London}
  \icmlaffiliation{ucla}{University of California, Los Angeles}
  \icmlaffiliation{nyu}{New York University}

  \icmlcorrespondingauthor{Yikai Zhang}{yikzhang@amd.com}

  \icmlkeywords{coding agents, reinforcement learning, training data, label errors, SWE-Gym}

  \anon{}{\vskip 0.13in
    \centerline{\releaselink{https://github.com/Arist12/SWE-Gym-Verified}{\faGithub}{Code}\hspace{2.2em}%
      \releaselink{https://huggingface.co/datasets/Arist12/SWE-Gym-Verified}{\faDatabase}{Dataset}\hspace{2.2em}%
      \releaselink{https://arist12.github.io/writing/swe-gym-verified/}{\faBlog}{Blog}}}

  \anon{\vskip 0.3in}{\vskip 0.2in}
]

\printAffiliationsAndNotice{}

\input{sections/abstract}
\input{sections/intro}
\input{sections/related}
\input{sections/prelim}
\input{sections/audit}
\input{sections/experiments}
\input{sections/discussion}

\bibliography{refs}
\bibliographystyle{icml2027}

\input{sections/ai_disclosure}

\newpage
\appendix
\onecolumn
\input{sections/appendix}

\end{document}

%% file: sections/abstract.tex
\begin{abstract}
Reinforcement learning (RL) trains coding agents on software tasks whose hidden tests decide the reward, yet task collections check these tests only against the reference fix. We ask what makes a task good training data, focusing on the correctness of its reward labels. We audit SWE-Gym with two LLM families, and a suspected error counts only if execution reproduces it. We find that unsolved often means mislabelled: 74\% of the tasks that a 9B base model never solves have a test that rejects correct fixes (a false negative), versus 20\% of those it sometimes solves. We repair or remove such tasks, and the audited tasks that remain train better policies. At equal size and matched difficulty, models trained on them score 2.4 and 2.7 points higher on held-out tasks from SWE-bench Verified and SWE-bench Pro than models trained on the original tasks, whereas simply dropping the never-solved tasks shows no such gain. We further find that wrong patches that pass (false positives) make up about a quarter of all passing patches, and that new tests that catch a quarter of them do not change held-out scores detectably. A pass rate shows whether a task yields reward, not whether that reward is right, so the reward labels themselves must be checked. To this end, we release \SGV with its 1{,}087 audited tasks, every verdict with its executed evidence, our training sets, and the full audit pipeline.
\end{abstract}

%% file: sections/intro.tex
\section{Introduction}
\label{sec:intro}

Reinforcement learning (RL) has become a standard way to train coding agents~\citep{luo2025deepswe,yang2026kimidev}, and task collections such as SWE-Gym, R2E-Gym, SWE-smith, and SWE-rebench supply the training tasks~\citep{pan2025swegym,jain2025r2egym,yang2025swesmith,badertdinov2025swerebench}. In collections mined from GitHub, a task pairs a repository and an issue with the tests of the pull request that resolved it~\citep{jimenez2024swebench}. The agent sees the issue but not the tests, and its patch is rewarded only if the tests pass. The hidden tests are therefore the reward labels. Yet a collection admits a task after checking only that the reference fix passes the tests and the original code fails them. This check shows that the reward is attainable. It does not show that every correct fix obtains it, or that only correct fixes do.

\begin{figure*}[t]
  \centering
  \includegraphics[width=\linewidth]{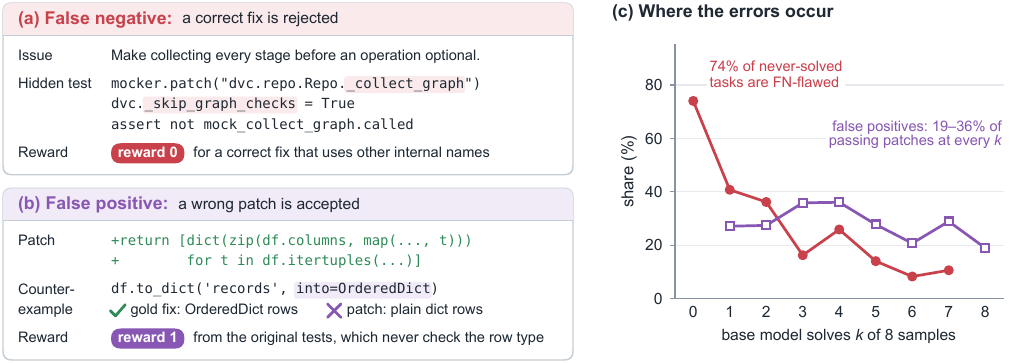}
  \caption{\textbf{Hidden tests mislabel patches in two ways, and the two error types occur in different places.} (a)~A false negative in \emph{iterative/dvc-3490}: the test relies on a private method and a private flag that the issue never names. (b)~A false positive in \emph{pandas-dev/pandas-50796}: a base-model patch passes every original test but ignores \texttt{into=}, which a counterexample exposes. (c)~Share of \Bset tasks that are FN-flawed and share of passing patches on \Aset tasks that are false positives, by the number $k$ of eight base-model samples that pass. For false positives, $k$ counts passes among the eight samples drawn for that measurement.}
  \label{fig:teaser}
\end{figure*}

Tests can mislabel a patch in two ways (Fig.~\ref{fig:teaser}a,\,b). A \emph{false negative} (FN) is a correct fix that the tests reject, for example because a test calls a private helper that the issue never mentions. A \emph{false positive} (FP) is a wrong patch that the tests accept, for example one that handles the reported case but breaks a nearby code path that no test covers. Both are well documented in evaluation benchmarks, where even human screening misses many of them~\citep{chowdhury2024swebenchverified,openai2026sweverified,wang2026patchdiff}. Training data receives coarser checks. RL recipes drop the tasks that the model never solves~\citep{yang2026kimidev} or the groups of samples without reward variance~\citep{yu2025dapo}, without looking at the tests. A pass rate tells whether a task yields reward, not whether that reward is right. LLM judges label each task's issue and tests only as a whole~\citep{oliva2025spice,badertdinov2025swerebench,badertdinov2026swerebenchv2}. What is missing is an execution-backed measurement of both error types in a real training set and a controlled test of their effect on RL.

We ask what makes a task good training data for a coding agent, and answer in two steps (Fig.~\ref{fig:pipeline}). First, we audit SWE-Gym~\citep{pan2025swegym} one fail-to-pass assertion at a time. Claude Opus~5.5 and GPT-6.1~Sol judge each assertion against the issue. A flagged assertion counts as an error only if the tests reject an executed alternative fix that follows the issue, and a passing patch counts as a false positive only if an executed counterexample exposes it. Tasks with a false negative are repaired by adding the missing requirement to the issue, or removed. The 1{,}087 remaining tasks form \SGV. Second, we train a 9B base model with GRPO on three sets of \Nset tasks with matched difficulty: original tasks (\Bset), audited tasks (\Aset), and audited tasks with added tests against false positives (\AFP). A fourth set, \BF, applies a pass-rate filter to the original tasks, dropping the never-solved ones. We evaluate every run on held-out tasks from SWE-bench Verified and SWE-bench Pro~\citep{deng2026swebenchpro}. We organize the study around four research questions:

\begin{itemize}[leftmargin=1.2em,itemsep=1.5pt,topsep=2pt]
  \item \textbf{RQ1: Where do label errors occur?} \emph{Unsolved often means mislabelled} (\S\ref{sec:rq1}). In \Bset, 74\% of the never-solved tasks have a test that rejects correct fixes, versus 20\% of the sometimes-solved ones. False positives make up about a quarter of all passing patches and are common at every difficulty level (Fig.~\ref{fig:teaser}c).
  \item \textbf{RQ2: Which label errors reach the training signal?} \emph{Pass-rate filters keep the label errors that train the policy} (\S\ref{sec:rq2}). GRPO learns from a task only if its samples' rewards differ. Never-solved flawed tasks almost never carry gradient, so dropping them changes little. Sometimes-solved flawed tasks contribute a fifth of the gradient-carrying groups, with or without the filter.
  \item \textbf{RQ3: Do audited tasks train better policies?} \emph{Yes, and a pass-rate filter does not substitute for the audit} (\S\ref{sec:rq3}). At equal size and matched difficulty, \Aset scores 2.4 points above \Bset on SWE-bench Verified and 2.7 on SWE-bench Pro over three seeds. Averaged over both benchmarks, \BF scores no higher than \Bset.
  \item \textbf{RQ4: Does hardening against false positives change the policy?} \emph{Not detectably} (\S\ref{sec:rq4}). New tests that catch a quarter of the false positives lower the training reward on the hardened tasks by 13 points, but re-grading \Aset's own patches with the same tests lowers its reward by about as much. Held-out pass@1 does not change detectably.
\end{itemize}

\noindent These answers indicate that a good training task is one whose tests reward exactly the correct fixes. Neither pass rate nor training reward can certify this. Of the factors that make data good, we study only the correctness of the reward labels, and control for size and difficulty. We release \SGV, every audit verdict with its evidence, the training sets, and the full audit pipeline (Appendix~\ref{app:repro}).

%% file: sections/related.tex
\section{Related Work}
\label{sec:related}

\paragraph{Label errors in software tasks.} Label errors are pervasive even in the test sets of standard benchmarks~\citep{northcutt2021labelerrors}. In supervised learning, the examples that a model finds hard to learn are often the mislabelled ones~\citep{swayamdipta2020cartography}. We find the same pattern in the reward labels of RL tasks. SWE-bench~\citep{jimenez2024swebench} accepts a task if the reference fix passes its tests and the original code fails them, and training collections apply the same check to thousands of tasks~\citep{pan2025swegym,jain2025r2egym,yang2025swesmith,badertdinov2025swerebench}. Agentic pipelines now build the environments across languages and operating systems, but they still take tasks from issues and the tests of their pull requests and admit them by the same check~\citep{li2026repolaunch}. This check cannot detect a test that rejects other correct fixes. Human screening has focused on evaluation sets and has proved hard to get right. SWE-bench Verified drops tasks whose tests may reject valid solutions~\mbox{\citep{chowdhury2024swebenchverified}}, yet a later audit finds such tests in most of the Verified tasks that models often fail~\citep{openai2026sweverified}. SWE-bench Pro's issues carry human-written requirements~\citep{deng2026swebenchpro}, but an audit estimates that about 30\% of its tasks are broken~\citep{openai2026swepro}. For training sets, PAIChecker's annotators label SWE-Gym's tasks for misalignment between the pull request and the issue~\citep{wang2026paichecker}. LLM judges~\citep{oliva2025spice,badertdinov2025swerebench,badertdinov2026swerebenchv2,li2026repolaunch} label each task's issue and tests as a whole. Such task-level labels can justify dropping a task, but they do not say which assertion is at fault. Tests that accept wrong programs are a known problem in code generation~\citep{liu2023evalplus}. Many patches accepted on SWE-bench are wrong~\citep{aleithan2024swebenchplus,yu2025utboost,wang2026patchdiff}, and strengthened tests reject a fifth of the patches that pass SWE-bench Verified~\citep{yu2026sweabs}. Coding agents also exploit tests that conflict with the task statement~\citep{zhong2026impossiblebench}.

\paragraph{Reward errors in RL.} Analyses of noisy verifiers assume that errors are random or of one fixed type. Under random noise, whether GRPO learns depends on the gap between the verifier's true- and false-positive rates~\citep{rad2026rateorfate}. With systematic errors, false negatives mainly delay learning, while false positives can stall or collapse it~\citep{egashira2026delay}. Recovering a verifier's false negatives improves GRPO on math~\citep{xu2026tinyv}. Even random rewards raise the math accuracy of some base models~\citep{shao2026spurious}, so a gain over the base model does not show that the labels are correct. We therefore compare training sets with each other. In competitive programming, where the statement fully specifies which outputs are correct, tests with fewer errors improve RL~\citep{wang2025codecontestsplus,he2026hardtestgen}. In repository-level tasks, false negatives arise from what the issue leaves unsaid, so we repair the statement instead of the tests. SWE-rebench V2 leaves the comparison of filtered and unfiltered training to future work~\citep{badertdinov2026swerebenchv2}.

\paragraph{Closest studies.} The Verification Horizon~\citep{wang2026verificationhorizon} rates whole tasks with an agentic judge, finds that never-solved tasks are more often of low quality, and reports gains from filtering. The concurrent Arctic-SWE~\citep{liu2026arcticswe} curates three collections with executed counterfactual patches, rewrites underspecified issues, and reports gains from curation and from stronger tests, in single runs. Unlike both, we judge individual assertions and release the executed evidence behind every verdict, so that each verdict can be checked. We break the errors down by whether they carry gradient, which shows why a pass-rate filter keeps the errors that train the policy, and we compare sets of equal size and matched difficulty over three seeds.

\begin{figure*}[t]
  \centering
  \includegraphics[width=\linewidth]{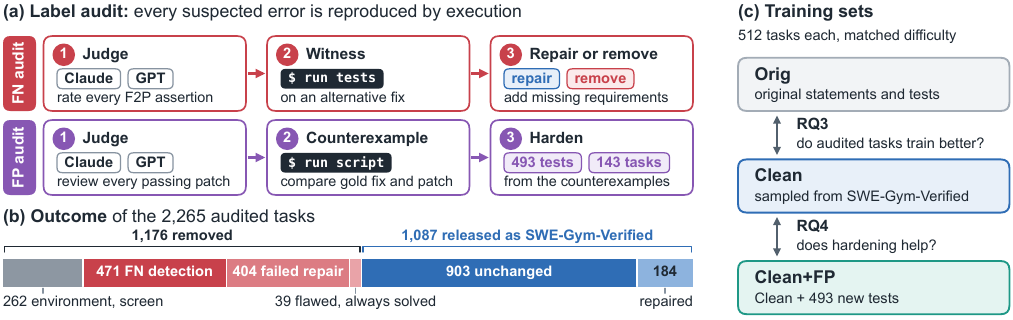}
  \caption{\textbf{From SWE-Gym to \SGV and three matched training sets.} (a)~The audit reproduces every suspected label error by execution: a flagged assertion must reject an executed witness fix, and a suspected wrong patch must fail an executed counterexample. (b)~Outcomes of the 2{,}265 tasks that enter the audit, in pipeline order (two unfinished audits are omitted). (c)~\Bset is sampled from screened original tasks. \Aset and \AFP are drawn from \SGV with \Bset's difficulty quotas (\S\ref{sec:setup}).}
  \label{fig:pipeline}
\end{figure*}

%% file: sections/prelim.tex
\section{Hidden Tests as Reward Labels}
\label{sec:defs}

\paragraph{Tasks and rewards.} A task $\tau$ consists of a repository snapshot, a statement $s$ (the issue, plus any requirements we add), hidden tests, and a reference fix (the gold patch). A fix must make the \emph{fail-to-pass} (F2P) tests pass without breaking the \emph{pass-to-pass} (P2P) tests, so a patch $p$ receives reward $R_\tau(p)=1$ if all tests pass after applying it, and $0$ otherwise. The agent sees $s$ and the repository, but not the tests. A task's difficulty is its \emph{pass count} $k$, the number of passes among eight base-model samples: a task is \emph{never solved} ($k=0$), \emph{sometimes solved} ($1\le k\le 7$), or \emph{always solved} ($k=8$). We call $p$ \emph{correct} if it resolves the problem as $s$ describes it and breaks no existing behaviour. The tests can mislabel $p$ in two ways. Appendix~\ref{app:terms} lists our terms.

\paragraph{False negatives.} A false negative is a correct patch with reward 0. We trace false negatives to individual \emph{assertions} of the F2P tests: an \texttt{assert}, a \texttt{pytest.raises} block, a call that must not fail, or an expected line of output. An assertion rejects correct patches when passing it requires information that neither $s$ nor the repository provides, such as an interface that the statement does not specify, an exact string, an arbitrary design choice, or behaviour beyond the issue. A task with such an assertion is either \emph{repaired}, if adding the missing requirement to $s$ removes the error, or \emph{FN-flawed}, if it does not. A test that requires the upstream wording of a message is a fair regression test but a wrong reward label for the task as stated, since it separates patches by information the agent never had. Details inferable from the repository, such as a naming convention, are not errors.

\paragraph{False positives.} A false positive is a wrong patch with reward 1. We count a passing patch $p$ as a false positive only if two LLMs label it incorrect and an executed \emph{counterexample} confirms the label. A counterexample is a script whose output is the same for the original code and the gold patch but differs for $p$, revealing a regression. This rule is conservative, because it misses incomplete fixes and wrong patches without a runnable counterexample.

\paragraph{How label errors reach GRPO.} GRPO computes each sample's advantage relative to the other samples of the same task, so a \emph{group} of samples \emph{carries gradient} only if its rewards are not all equal. Label errors can therefore affect training in three ways. (i)~If the tests reject every reasonable patch, all samples receive reward~0, and the task consumes compute without yielding gradient~\citep{xu2026tinyv}. (ii)~If the tests reject only some reasonable patches, the reward separates correct patches by details that the statement never specifies, and the gradient favours the patches that happen to match them. (iii)~A false positive rewards a wrong patch, which can stall or derail learning~\citep{egashira2026delay}.

%% file: sections/audit.tex
\section{SWE-Gym-Verified: An Execution-Backed Audit of SWE-Gym}
\label{sec:pipeline}

We audit the 2{,}265 usable, de-duplicated tasks among SWE-Gym's 2{,}438 with Claude Opus~5.5 and GPT-6.1~Sol (Fig.~\ref{fig:pipeline}). Each model and its coding agent form a \emph{family}, Claude or GPT (Appendix~\ref{app:pipeline}). After screening each task's environment and statement, the audit follows one principle: no error verdict rests on judgment alone. Every suspected error must be reproduced by execution. The tests must reject a concrete alternative fix, or a counterexample must expose a concrete patch. Whether the alternative is reasonable remains an LLM judgment, made by the family that wrote it and, when the two families disagree, by both.

\subsection{False Negatives: Judge Every Assertion, Then Execute a Witness}
\label{sec:fn}

Both families rate every F2P assertion against the statement. They see the test code, the gold patch, and the relevant repository code, but never a task's pass rate or difficulty (Appendix~\ref{app:pipeline}). An assertion is \emph{flagged} if either family judges that it may reject a correct patch, and unflagged assertions are cleared without execution. For every flagged assertion, each family's coding agent writes a \emph{witness}, a patch that follows the statement but differs from the gold patch in the flagged detail. The agent then runs the tests on the witness in the task container. A task receives an \emph{FN verdict} when one of its flagged assertions rejects a witness that the authoring family considers a reasonable reading of the statement. On 374 tasks, in most cases because the families' ratings disagree or are unsure, each family reviews the other's witness and the two debate, which clears 171 of them. Of the 1{,}091 tasks with an FN verdict, 963 are \emph{confirmed} by both families and 128 are \emph{contested}, upheld by only one. In 810 of the confirmed verdicts, witnesses from both families fail the same test (Table~\ref{tab:tiers}, Appendix~\ref{app:audit}).

\paragraph{Repair.} We repair a task with an FN verdict rather than drop it, unless no change of the statement can repair it or the base model always solves it. As in SWE-bench Pro~\citep{deng2026swebenchpro}, a repair appends a \emph{Requirements} block that states the missing requirement without revealing the solution. The F2P tests stay unchanged (one task has a relaxed test instead). A repair is accepted only if Claude Opus~5.5 and GPT-5.6~Sol, a second GPT model, find no leak (Appendix~\ref{app:pipeline}) and the repaired task passes the audit again (184 of 588 attempts). Otherwise the task is removed as FN-flawed.

\subsection{False Positives: Confirm by Counterexample,\texorpdfstring{\newline}{ }Then Harden}
\label{sec:fp}

On \Aset's tasks, we collect the distinct passing patches among eight base-model samples. Both families label each patch without seeing the gold patch or the tests, and we count false positives as defined in \S\ref{sec:defs}. For \AFP, we \emph{harden} the tests by turning the confirming counterexamples into new tests. A test is kept only if it passes with the gold patch and with every patch not labelled a false positive, which leaves 493 tests on 143 tasks (Appendix~\ref{app:fp}).

\subsection{The Release and the Quality of Its Verdicts}
\label{sec:release}

\SGV contains 1{,}087 tasks, 184 of them with a repaired statement. \emph{Verified} means that two LLM families rated every F2P assertion and that flagged assertions were tested by execution. Humans checked a sample of the verdicts, not every task as in SWE-bench Verified. Each task records its pass count on its final statement and tests: 138 released tasks are never solved, 570 sometimes solved, and 379 always solved.

\paragraph{Verdict quality.} Table~\ref{tab:quality} summarizes four checks of the verdicts (Appendix~\ref{app:audit}).

\begin{table}[t]
\centering
\caption{\textbf{Checks of the audit verdicts.} Humans: two of the authors, blind to the verdicts (``3, 1'': one count per annotator). Open: not determined by the statement or the repository. PAIChecker-marked: its annotators mark tests or patches that go beyond the issue~\citep{wang2026paichecker}.}
\label{tab:quality}
\footnotesize
\setlength{\tabcolsep}{5pt}
\begin{tabular}{@{}lrr@{}}
\toprule
Check & Items & Result \\
\midrule
Claude and GPT flag the same tasks & 2{,}001 & $\kappa=$~0.80 \\
\addlinespace[2pt]
Both humans: flagged assertion open & 45 & 33 \\
\quad never / sometimes solved & 8 / 37 & 8 / 25 \\
Each human: unflagged control open & 30 & 3, 1 \\
Each human: false positive incorrect & 15 & 10, 12 \\
\addlinespace[2pt]
FN verdict, PAIChecker-marked tasks & 726 & 82\% \\
FN verdict, other tasks & 1{,}149 & 37\% \\
\addlinespace[2pt]
FN in a random sample of cleared tasks & 79 & 2 \\
\bottomrule
\end{tabular}
\end{table}

%% file: sections/experiments.tex
\section{Experiments}
\label{sec:experiments}

\subsection{Setup}
\label{sec:setup}

\paragraph{Training sets.} We compare three matched training sets of \Nset tasks (Fig.~\ref{fig:pipeline}c). \Bset is a sample of screened SWE-Gym tasks with their original statements and tests, stratified by difficulty. \Aset is drawn from \SGV with the same random ranking of tasks as \Bset and \Bset's quota of tasks in each difficulty range. \Aset and \Bset therefore match in size and in base-model pass rate (49.8\% and 48.3\%). They share 314 tasks. \AFP is \Aset plus the new tests. \Bset is a strong baseline. Its sampling pool already excludes unusable and duplicate tasks, pull requests labelled defective, and statements rated underspecified. In addition, \Bset's share of never-solved tasks is fixed at 15\%, versus 54\% in the pool. Only 42 \Aset tasks have a repaired statement, so comparing \Aset with \Bset mainly tests the removal and replacement of flawed tasks (Appendix~\ref{app:sets}). A fourth set, \BF, applies a pass-rate filter to \Bset: it replaces \Bset's 77 never-solved tasks with further sometimes-solved tasks from the same pool. It is easier than the other sets (base-model pass rate 56.8\%), and 97 of its tasks are FN-flawed.

\paragraph{Training and evaluation.} We train the base model, Qwen3.5-9B~\citep{qwen2026qwen35}, with GRPO~\citep{shao2024deepseekmath} in the SWE-agent scaffold~\citep{yang2024sweagent}. Each run consists of 80 steps of 16 tasks with 8 samples each (2.5 passes over a set), and every set uses the same three seeds (Table~\ref{tab:settings}, Appendix~\ref{app:settings}). We report pass@1, the fraction of attempts that resolve a task, on two held-out benchmarks that share no repository with the training tasks. \textbf{\mbox{Verified-397}} consists of the 397 scored tasks of SWE-bench Verified outside our 100-task development split, \textbf{dev-100}. For \textbf{SWE-bench Pro}, we use the public set of version~2~\citep{dan2026swebenchprov2}, which also differs from training in languages, scaffold, and budget. For each run we report the checkpoint with the best dev-100 pass@1. Results are means over three seeds, with differences in percentage points (pp).

\begin{figure*}[t]
  \centering
  \includegraphics[width=\linewidth]{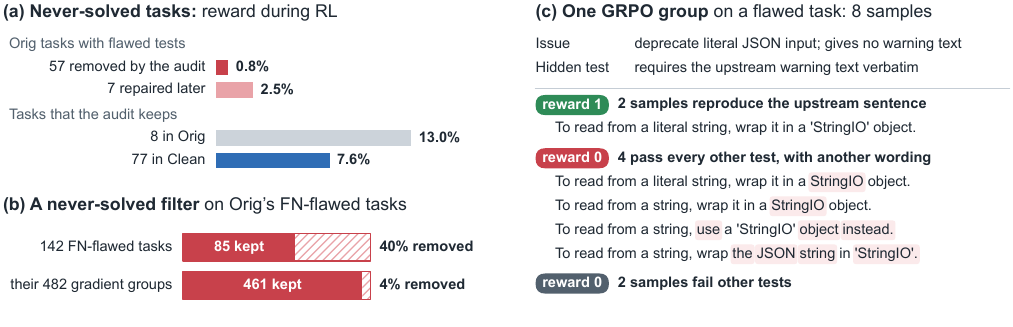}
  \caption{\textbf{Never-solved flawed tasks are silent, but the flawed tasks that survive a pass-rate filter train the policy.} (a)~Training reward on never-solved tasks by audit outcome (mean over seeds). Tasks that the audit finds flawed stay unsolved, while tasks that it keeps receive reward. The five never-solved \Bset tasks removed for other reasons are not shown. (b)~\Bset's 142 FN-flawed tasks and their 482 gradient-carrying GRPO groups, split by whether a filter that drops never-solved tasks keeps them or removes them (hatched). The flawed tasks that the filter keeps are 16.6\% of \Bset's tasks and receive 19.5\% of its gradient-carrying groups. (c)~One GRPO group of \emph{pandas-dev/pandas-53409}: the second sentence of the warning that each sample adds. Tinted words differ from the upstream sentence.}
  \label{fig:mechanism}
\end{figure*}

\subsection{RQ1: Where Do Label Errors Occur?}
\label{sec:rq1}

\paragraph{False negatives are common and concentrate on never-solved tasks.} Of the 2{,}001 audited tasks whose assertions were judged, 1{,}091 (55\%) receive an FN verdict. About half of these verdicts involve behaviour that the issue does not ask for. Interfaces that the statement does not specify, exact strings, and arbitrary design choices each appear in about a fifth (a verdict can have several causes). The errors are far from uniform across difficulty. In \Bset, 74\% of the never-solved tasks are FN-flawed, versus 20\% of the sometimes-solved ones, and the share falls as the pass count rises (Fig.~\ref{fig:teaser}c). The contrast does not hinge on the LLM verdicts being exact. It holds with only the verdicts confirmed by both families (68\% versus 13\%) and under PAIChecker's human labels of \Bset's pool (61\% versus 21\%; \citealp{wang2026paichecker}). It also holds when the tasks that we repair later count as flawed (83\% versus 28\%), with a larger sampling budget, and among tasks with the same number of F2P tests (Table~\ref{tab:contrast}, Appendix~\ref{app:audit}). Repairs test the verdicts directly. Once the missing requirement is stated, the base model's pass count on the repaired tasks rises by 2.1 out of eight on average, versus 0.0 on unchanged tasks that are re-measured. One appended sentence takes \emph{getmoto/moto-4846} from zero to seven of eight (Appendix~\ref{app:cases}).

\paragraph{False positives are common at every difficulty level.} Of the 1{,}854 distinct patches that pass on \Aset's tasks, 26.1\% are false positives under our conservative rule, and another 37.3\% are undetermined (not labelled correct by both families). At every pass count, false positives are 19--36\% of the passing patches (Fig.~\ref{fig:teaser}c), and they are spread across tasks: 224 of the 432 tasks with a passing patch have at least one.

\takeaway{1}{Unsolved often means mislabelled.}{A never-solved task is more likely than not to have a test that rejects correct fixes, so a zero pass rate cannot be read as hardness alone. False positives are common at every difficulty level.}

\subsection{RQ2: Which Label Errors Reach\texorpdfstring{\newline}{ }the Training Signal?}
\label{sec:rq2}

We trace \Bset's FN-flawed tasks through its three GRPO runs: 3{,}840 groups of eight samples, 2{,}365 of which carry gradient (Fig.~\ref{fig:mechanism}, Appendix~\ref{app:budget}).

\paragraph{Never-solved flawed tasks are silent.} The 57 never-solved FN-flawed tasks make up 11.1\% of \Bset's groups but only 0.9\% of its gradient-carrying groups. Training does not rescue them. Their training reward stays at 0.8\%, while the never-solved tasks that the audit keeps do receive reward (7.6\% in \Aset, Fig.~\ref{fig:mechanism}a). The FN raters never saw a pass rate, yet their verdicts separate the never-solved tasks that receive no reward in training from those that do. The gradient that such a task does carry can point the wrong way. In \emph{getmoto/moto-5137}, the test also requires a second API that the issue never mentions. Of 40 training samples, 30 do what the issue asks but get no reward, and the only rewarded sample goes beyond the issue (Appendix~\ref{app:cases}).

\paragraph{Sometimes-solved flawed tasks train the policy.} The 85 sometimes-solved FN-flawed tasks make up 16.6\% of \Bset's tasks and contribute 19.5\% of its gradient-carrying groups (12.2\% counting only verdicts confirmed by both families). To see what the flagged assertions do in these groups, we mask them in the tests and re-run the tests on the zero-reward samples of the two runs with test-level records. On the 20 of these tasks where every flagged assertion can be masked, 30\% of the zero-reward samples then pass, and 26 of the 49 gradient-carrying groups contain such a sample (Appendix~\ref{app:budget}).

\paragraph{A pass-rate filter removes only the silent errors.} Dropping never-solved tasks~\citep{yang2026kimidev} would discard 40\% of \Bset's FN-flawed tasks, which looks effective. But it removes none of the sometimes-solved ones, and so only 4\% of the gradient-carrying groups on FN-flawed tasks (Fig.~\ref{fig:mechanism}b). Filters on pass rate or on reward variance~\citep{yu2025dapo} see whether a task is solved, not why it fails. Training on \BF confirms this. All-zero groups fall from 20.7\% to 8.3\%, yet sometimes-solved FN-flawed tasks still contribute 19.9\% of the gradient-carrying groups.

\paragraph{What the surviving errors reward.} Fig.~\ref{fig:mechanism}c shows one sometimes-solved FN-flawed task. The issue asks to deprecate literal JSON input and gives no warning text, but the hidden test requires the upstream text verbatim. In one GRPO group, every sample adds the warning. The two that pass reproduce the upstream sentence, and four others fail only this test, so the GRPO advantage favours reproducing the upstream text over fixing the issue (Appendix~\ref{app:cases}).

\takeaway{2}{Pass-rate filters keep the errors that train the policy.}{Dropping never-solved tasks removes label errors that carry almost no gradient. The flawed tasks that do carry gradient are found only by checking the labels.}

\begin{figure*}[t]
  \centering
  \includegraphics[width=\linewidth]{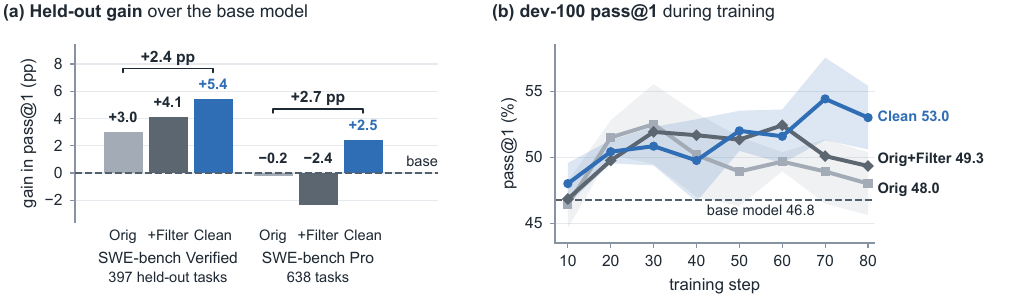}
  \caption{\textbf{Audited data scores higher on both held-out benchmarks, whereas a pass-rate filter shows no such gain.} (a)~Gain in pass@1 over the base model at each run's selected checkpoint, mean over seeds (+Filter: \BF). Brackets give \Aset $-$ \Bset. (b)~dev-100 pass@1 at every checkpoint (lines: mean over seeds; shading: $\pm$ one standard deviation for \Bset and \Aset; labels: mean at step~80).}
  \label{fig:curves}
\end{figure*}

\begin{figure*}[t]
  \centering
  \includegraphics[width=\linewidth]{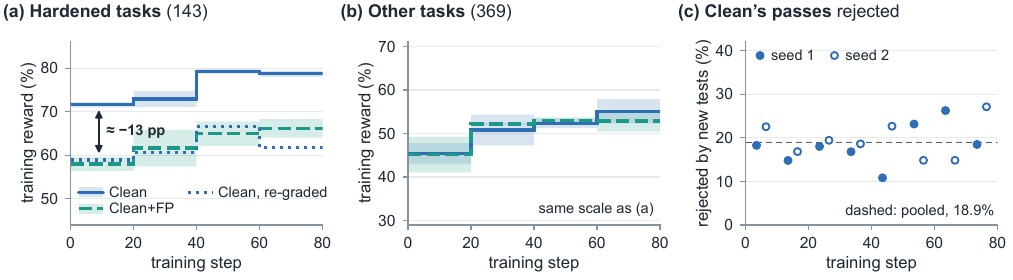}
  \caption{\textbf{Hardening lowers the reward where tests were added, about as much as re-grading predicts.} (a)~Training reward on the 143 hardened tasks in 20-step bins (mean $\pm$ one standard deviation over seeds). Re-graded: \Aset's patches under the new tests (two seeds). (b)~The same for the 369 other tasks. (c)~Share of \Aset's passing patches on the hardened tasks that the new tests reject, by the ten-step bin in which a patch first appears.}
  \label{fig:fp}
\end{figure*}

\subsection{RQ3: Do Audited Tasks Train Better Policies?}
\label{sec:rq3}

\paragraph{Audited data scores higher on both held-out benchmarks.} \Aset outperforms \Bset by 2.4\pp on \Vfour and by 2.7\pp on SWE-bench Pro (Table~\ref{tab:heldout}, Fig.~\ref{fig:curves}a). On Pro, \Bset does not improve over the base model ($-0.2$\pp). The runs of \Aset and \Bset overlap on each benchmark, but averaged over the two benchmarks every \Aset run gains more than every \Bset run (Appendix~\ref{app:results}). On each benchmark, \Aset $-$ \Bset remains at least 1.4\pp when any single run is left out. A pass-rate filter yields no such gain, although \BF's tasks are easier. \BF is 1.1\pp above \Bset on \Vfour and 2.1\pp below it on Pro. On average over both, \BF gains $+0.9$\pp over the base model, versus $+1.4$\pp for \Bset and $+3.9$\pp for \Aset.

\begin{table}[t]
\centering
\caption{\textbf{Held-out pass@1 (\%)} at each run's selected checkpoint: mean $\pm$ standard deviation over three seeds. Gain: each run's gain over the base model, averaged over the two benchmarks (pp).}
\label{tab:heldout}
\small
\begin{tabular}{@{}lccc@{}}
\toprule
Training set & \Vfour & SWE-bench Pro & Gain \\
\midrule
Base model & 57.2 & 53.8 & -- \\
\Bset & 60.2\pmstd{2.1} & 53.5\pmstd{2.1} & $+$1.4\pmstd{0.9} \\
\BF & 61.3\pmstd{2.1} & 51.4\pmstd{3.0} & $+$0.9\pmstd{2.5} \\
\Aset & 62.6\pmstd{1.1} & 56.2\pmstd{1.3} & $+$3.9\pmstd{1.2} \\
\AFP & 62.6\pmstd{1.0} & 57.3\pmstd{3.7} & $+$4.5\pmstd{2.2} \\
\bottomrule
\end{tabular}
\end{table}

\paragraph{The gap emerges late in training.} On dev-100, where every run is evaluated at the same steps, the sets stay close through step~40 (Fig.~\ref{fig:curves}b). \Bset peaks at step~30 (52.5\%) and falls to 48.0\% by step~80, while \Aset ends at 53.0\% and \BF at 49.3\%. At step~80, every \Aset run scores higher than every \Bset run (Appendix~\ref{app:results}). Checkpoint selection therefore picks \Bset at steps 20--30, near its peak, and \Aset at 60--70 (Table~\ref{tab:main}).

\paragraph{What explains the gain.} The gain does not come from more gradient-carrying groups. \BF has the most (70.8\%, versus 65.9\% for \Aset and 61.6\% for \Bset), yet it gains no more than \Bset. Among the remaining differences (Table~\ref{tab:sets}), two follow from the audit. First, a fifth of \Bset's and of \BF's gradient-carrying groups fall on tasks whose tests reject correct fixes (\S\ref{sec:rq2}). Second, \Aset's never-solved tasks are audited tasks with attainable reward. They contribute 6.9\% of its gradient-carrying groups, versus 2.6\% for \Bset's never-solved tasks and none in \BF. Our runs cannot disentangle the two.

\takeaway{3}{Audited tasks train better policies.}{At equal size and matched difficulty, audited tasks score higher on both held-out benchmarks. A pass-rate filter shows no such gain, so in our runs it does not substitute for auditing the labels.}

\subsection{RQ4: Does Hardening Against False Positives Change the Policy?}
\label{sec:rq4}

We train on \AFP with \Aset's seeds and task order. The new tests target regressions, cover 143 of the \Nset tasks, and catch 25\% of the false positives in a sample of passing training patches (Appendix~\ref{app:rq4}).

\paragraph{The policy barely adapts to the new tests.} On the hardened tasks, \AFP's training reward is about 13\pp below \Aset's throughout training, while the reward on other tasks stays close to \Aset's (Fig.~\ref{fig:fp}a,\,b). A drop is expected even if the policy does not change, since the new tests reject patches that the old ones accept. We therefore re-grade the passing patches of the \Aset runs, which never saw the new tests. Re-grading lowers \Aset's reward by about as much, so averaged over training the \AFP policy passes the new tests only 0.4 and 1.8\pp more often than the \Aset policy would (two seeds). Nor do false positives fade. The new tests reject about 19\% of \Aset's passing patches on the hardened tasks throughout training (Fig.~\ref{fig:fp}c).

\paragraph{Held-out pass@1 does not change detectably, and cannot show fewer false positives.} \AFP $-$ \Aset is $-0.1$\pp on \Vfour and $+1.1$\pp on Pro (Table~\ref{tab:heldout}), well within seed-to-seed variation, but three seeds cannot establish that the two sets are equivalent. Held-out pass@1 is graded by the benchmarks' original tests, which accept false positives, so it cannot reveal whether the policy writes fewer wrong patches that pass. The concurrent Arctic-SWE reports a held-out gain from strengthening tests, in single runs on 4{,}889 tasks whose tests it had found too loose~\citep{liu2026arcticswe}, whereas our new tests cover about a quarter of \AFP's tasks.

\takeaway{4}{Training reward cannot rank datasets.}{New tests that catch a quarter of the false positives lower the training reward, but the \AFP policy passes them only 0.4 and 1.8\pp more often than the \Aset policy would. Held-out pass@1 does not change detectably. Training reward is computed with the labels under study, so it cannot show which labels are better.}

%% file: sections/discussion.tex
\section{Discussion: What Makes Good\texorpdfstring{\newline}{ }Training Data?}
\label{sec:discussion}

For reward labels, the four answers yield three criteria. \textbf{Every correct fix must obtain the reward.} Most tasks that the base model never solves have a test that rejects correct fixes, so filters and curricula that read a zero pass rate as hardness act largely on label errors. \textbf{The reward must vary for the right reason.} The label errors that train the policy lie in tasks whose samples receive different rewards, so filters on pass rate or on reward variance~\citep{yu2025dapo} cannot detect them. \textbf{Quality must be measured on held-out tasks,} at matched size and difficulty. \BF attains the highest training reward of all sets (Table~\ref{tab:sets}) but shows no held-out gain over \Bset on average. \Aset's training reward exceeds \Bset's partly because flawed tests cap \Bset's reward. \AFP attains a lower training reward than \Aset (Fig.~\ref{fig:fp}a) but shows no detectable held-out difference.

Two questions stand out. The first is how to remove false positives. Counterexample tests catch a quarter of them, and our 484 false positives with executed counterexamples can serve as a benchmark for verifiers that go beyond tests. The second is how cheap an audit can be. Without execution, Claude's rating alone matches the final verdict on 91\% of the tasks and GPT's on 95\% (Appendix~\ref{app:audit}), but whether one such pass suffices for RL is untested.

\paragraph{Limitations.} We study one base model, scaffold, RL recipe, and task source, with three seeds per set and an 80-step budget. A task's difficulty is relative to this model and its sampling budget. \Aset and \Bset differ in the tasks that the audit removes, repairs, and replaces, so RQ3 measures these effects jointly. The verdicts are LLM judgments backed by execution and checked on 105 items by two of the authors (Appendix~\ref{app:audit}). FN detection covers only F2P assertions, and the counts of false positives cover only regressions. The held-out benchmarks can still contain label errors of both kinds~\citep{openai2026sweverified,dan2026swebenchprov2}. We assume that these errors affect all training sets alike.

\section{Conclusion}
\label{sec:conclusion}

In RL for coding agents, the hidden tests label every patch, and in SWE-Gym these labels are often wrong. Of the tasks that we judged, 55\% have a test that rejects correct fixes. The errors that a pass-rate filter keeps are the ones that carry gradient. Models trained on audited tasks score 2.4 and 2.7 points higher on the two held-out benchmarks than models trained on the original tasks, whereas filtering the original tasks by pass rate shows no such gain. It remains open whether these findings hold at larger scale and for other models, whether repair outperforms removal, and how to remove false positives. \SGV, its audit records, our training sets, and the audit pipeline support work on these questions.

\section*{Impact Statement}

This work aims to make RL data for coding agents more reliable. Removing FN-flawed tasks avoids spending rollouts on tasks whose tests reject correct patches, and hardening tests against false positives reduces the reward that a policy can obtain with patches that only look correct, a form of reward hacking. Releasing per-task audit records lets others check and contest individual verdicts instead of \mbox{trusting} aggregate filters. Like other work on training coding agents, ours carries the usual dual-use concerns: more capable agents can write harmful code or automate work with economic consequences for developers. Our tasks come from public open-source repositories and their public issues, as in SWE-Gym, and we add no personal data. The audit relies on LLM judges whose biases we cannot fully characterize, so we release every verdict with its evidence and report contested cases neutrally (Appendix~\ref{app:cases}).

%% file: sections/ai_disclosure.tex
\section*{AI Disclosure Statement}

\textbf{Generative AI in the method.} LLMs are part of the audit that this paper reports, and \S\ref{sec:pipeline} and Appendix~\ref{app:pipeline} describe their use and give the model versions. Models of two families, Claude and GPT, screen statements, rate every F2P assertion, write and run witness patches, review each other's witnesses, write and check statement repairs, label passing patches, and write the counterexamples that become new tests; a separate LLM reviewer agent audited a sample of the rejections of these tests (Appendix~\ref{app:pipeline}). We do not take these outputs on trust: the audit backs its verdicts with executed witnesses and counterexamples, two of the authors labelled a blind sample of verdicts without AI assistance (\S\ref{sec:release}, Appendix~\ref{app:audit}), and we release the final version of every prompt and every verdict with its executed evidence.

\textbf{Generative AI in preparing the paper.} We conceived the study, chose its research questions, designed the audit and the experiments, and interpreted the results. We used generative AI tools for three tasks: to edit our drafts of the manuscript for wording; to write and restyle the plotting scripts of the figures, which plot numbers from our experimental records; and to review drafts critically, after which we decided what to change.

%% file: sections/appendix.tex
\section*{Appendix}
\addcontentsline{toc}{section}{Appendix}

\section{Released Artifacts}
\label{app:repro}

We release the data of this study in five parts.
\begin{itemize}[leftmargin=2.0em,itemsep=0pt,topsep=2pt]
  \item[(i)] \SGV: every released task with its final statement (repaired statements marked), its graded F2P and P2P tests, its difficulty for this base model and budget, its evidence tier (Table~\ref{tab:tiers}), and its audit flags.
  \item[(ii)] Per-task audit records for every task with a finished audit: both families' verdicts, witness patches with their execution outcomes, removal reasons, and repairs.
  \item[(iii)] The four training sets, with the sampling frames and samplers that reproduce \Bset, \Aset, and \BF, and the \AFP test patches with their construction records.
  \item[(iv)] The outcome of every held-out attempt of every reported run, with dev-100 at all eight checkpoints.
  \item[(v)] The code of the full audit pipeline with the final version of every prompt, the training and evaluation configurations, and a datasheet.
\end{itemize}
\anon{An anonymized copy of the release accompanies this submission as supplementary material.}{The dataset is at \url{https://huggingface.co/datasets/Arist12/SWE-Gym-Verified}, the code and the experiment files at \url{https://github.com/Arist12/SWE-Gym-Verified}, and a blog post with the main results and case studies at \url{https://arist12.github.io/writing/swe-gym-verified/}.}
The release flags 53 tasks whose issue contains the full solution or whose F2P tests appear unrelated to the issue; a strict subset without them has 1{,}034 tasks, including 495 of \Aset's \Nset. The data is meant for training and for studying reward errors, not for evaluation. Task records are versioned so that a verdict can be corrected when others dispute it. Our additions and our code are released under the Apache License 2.0; issue text and repository code keep their original terms.

\section{Audit Pipeline Details}
\label{app:pipeline}

\paragraph{Stages.} The pre-screen drops duplicates and tasks with unusable images, title-only statements, or execution failures. The environment stage runs the gold and empty patches offline and removes tasks whose gold patch fails, whose tests need the network, or whose F2P tests are flaky; it also truncates each container's git history to the task's base commit, so that a fix cannot be read from later history. The screening stage removes empty or link-only statements, pull requests that PAIChecker's human annotations~\citep{wang2026paichecker} label as defective, and statements that both families find misleading relative to the tests. A further screen flags tasks whose issue contains the full solution, and these tasks stay in the release (Appendix~\ref{app:repro}). For the 140 tasks without a PAIChecker label, Claude Opus~5.5 makes this judgment and removes six. FN-flawed tasks that the base model always solves (\band{8}) are removed without a repair attempt, because always-solved tasks are not used for training; always-solved tasks without an FN verdict stay in the release. Fig.~\ref{fig:pipeline}b gives the removals by stage. FN detection removes 471 tasks that no change of the statement can repair: most of them (310) by both families' judgment, and nearly all others because the test needs a private name or an internal module, or checks a mock's calls. Of the 404 tasks removed at repair, 341 failed one of the repair checks or the blind attempt (Repair, below), 32 failed the re-audit, and for 31 no repair could be written.

\paragraph{Models and interfaces.} The audit uses Claude Opus~5.5 and GPT-6.1~Sol. The GPT family has a second model, \mbox{GPT-5.6~Sol}. It is one of the two leak judges of the repair checks and rates statement quality for \Bset's pool (Appendix~\ref{app:sets}). The API identifiers are \texttt{claude-opus-5-5}, \texttt{gpt-6.1-sol}, and \texttt{gpt-5.6-sol}, without a snapshot date. Table~\ref{tab:models} gives the models and the interface of each step. It leaves out the reviewer agent of the hardening step (below), which ran as a separate agent session. The screens, the ratings, the P2P judgments, the repair writer, the mapping and leak checks, and the FP labels are API calls. Claude is called through the Anthropic Messages API with adaptive thinking, and the GPT models through the \mbox{OpenAI} Responses API. Every call asks for reasoning effort \emph{high}, sets no sampling parameter, and allows 32{,}000 output tokens (8{,}000 in the statement-quality filter). Witnesses, cross-reviews, debates, the judgment whether a statement change can repair a task, and the blind attempts at repaired tasks (Repair, below) are agent runs. Claude Opus~5.5 runs in Claude~Code~2.1.295 and GPT-6.1~Sol in Codex~CLI~0.156.1, both at reasoning effort \emph{high}. The agent process runs outside the task container. Each of its commands runs in a fresh container of the task image without network access, with two CPUs, 8\,GB of memory, and a limit of 900\,s. A blind attempt keeps one such container for the whole run. A run that executes a command anywhere else is discarded and repeated once. A witness run may take 90 minutes, a review or a debate 45, and a repair judgment 60, within a time budget per task (two hours for most tasks). All calls and agent runs reach the models through an API gateway of our institution. The statement-quality filter ran on 6 and 7 October 2026, the audit from 13 to 15 October, and the FP labelling from 19 to 21 October (UTC).

\begin{table}[h]
\centering
\caption{\textbf{Models and interfaces of the audit.} Both: Claude Opus~5.5 and GPT-6.1~Sol. Agent: Claude~Code or Codex~CLI.}
\label{tab:models}
\footnotesize
\begin{tabular}{@{}lll@{}}
\toprule
Step & Models & Interface \\
\midrule
Screen for empty or link-only statements; defective-PR check of unlabelled tasks & Claude Opus~5.5 & API \\
Solution-in-issue and misleading-statement screens & both & API \\
FN rating of every F2P assertion & both & API \\
Witness, cross-review, debate & both & agent \\
Judgment of a P2P test that rejects a patch passing all F2P tests & both & API \\
Judgment whether a statement change can repair a task & both & agent \\
Repair writer & Claude Opus~5.5 & API \\
Repair check: requirement mapping & both & API \\
Repair check: leak judges & Claude Opus~5.5, GPT-5.6~Sol & API \\
Blind attempt at a repaired task (only in part of the audit) & Claude Opus~5.5 & agent \\
FP labels with counterexample scripts & both & API \\
Statement-quality filter of \Bset's pool (Appendix~\ref{app:sets}) & GPT-5.6~Sol & API \\
\bottomrule
\end{tabular}
\end{table}

\paragraph{FN detection.} Both families rate every F2P assertion. They see the statement, the issue discussion (marked as not visible to the task-solving agent), the test source, the gold patch (marked as one possible solution), and a code-evidence pack built mechanically at the base commit. The pack shows the code that the gold patch edits, with its callers, callees, and similar code elsewhere; the repository definitions that the F2P tests use; and, for each identifier in the tests or the gold patch that the issue does not mention, whether it exists in the repository. Witness agents see the statement, the test patch, the F2P and P2P test names, and the gold patch as a reference. The FN raters and the witness agents see no pass rate, difficulty level, or training trajectory. A task's difficulty is used only later, in the choice between repair and removal, in the repair checks, and in the choice of cleared tasks that are escalated afterwards (Table~\ref{tab:tiers}).

Each family scores the task: 0 if no assertion can reject a correct patch, 1 if unsure, and 2 or more if it names a reasonable alternative patch that a listed assertion would reject. If both score 0, the task is cleared without execution. If both score 2 or more (\emph{witness mode}), each family's agent writes and runs its own witness. This mode has no cross-review, so each family judges only its own witness; in 810 of the 878 verdicts confirmed in this mode, the two witnesses fail a common F2P test (Table~\ref{tab:tiers}). In all other cases (\emph{full escalation}), each family reviews the other's executed witness, and they debate once. A flagged assertion that already passes on the original code also sends its task to full escalation. The verdict is then \emph{confirmed} if both families uphold it and \emph{contested} if only one does; a task with either verdict is removed or repaired. Outside this rating, both families judge a P2P test when it rejects a patch that passes every F2P test. In three tasks, they find P2P tests that require the old behaviour.

\paragraph{Repair.} A flagged task goes to repair unless a mechanical rule or both families' judgment finds that no change of the statement can repair it; one such rule is that the assertion names a private symbol. The repair writer (Claude Opus~5.5) appends a few public requirements that make every flagged assertion follow from the statement, without file paths, private names, or the solution. A repair must pass four checks. First, a mechanical check confirms that the repair has no paths or private names and that its quotations are verbatim. Second, both families check that each requirement maps to a flagged assertion. Third, two leak judges, Claude Opus~5.5 and \mbox{GPT-5.6~Sol}, read the repair, and a flag from either fails it (\mbox{GPT-5.6~Sol} caught more leaks than \mbox{GPT-6.1~Sol} in a pilot). Fourth, if two or more base-model samples on the repaired task fail the same assertion in the same way, one of them is escalated to cross-review and debate, and the task is removed if both families confirm the error. For part of the audit, Claude Opus~5.5 also attempted the repaired task without seeing the tests (a \emph{blind attempt}), and two tasks were removed because an attempt still failed a flagged assertion. The repaired task is then re-audited from scratch. On the 154 repaired tasks with a difficulty measurement on both statements, the base model's pass count rises by 2.1 out of eight; on 92 unchanged tasks that are re-measured, it changes by 0.0. Appendix~\ref{app:cases} shows two repaired tasks.

\paragraph{FP measurement and hardening.}
\label{app:fp}
Both families label each passing patch \emph{correct}, \emph{incorrect}, or \emph{uncertain}. They see the statement, a code pack, and the patch, but not the gold patch or the tests. An \emph{incorrect} label comes with counterexample scripts, which we run offline on the original code, the gold patch, and the candidate patch. Patches that are neither false positives nor labelled correct by both families are \emph{undetermined}; they include the 172 patches that both families label incorrect but that lack an executed counterexample.

Each confirmed counterexample becomes a candidate P2P test. It is kept only if it passes on the original code and the gold patch, fails on its FP patch, and passes on every passing patch of the task that is not a false positive. A task needs at least one such patch besides the gold patch. We then split the patches that pass during \Aset's training in half per task. On the \emph{construction half}, we delete every test that rejects a patch not labelled incorrect by both families. An LLM reviewer agent audited a seeded sample of the remaining rejections on that half and found 30 of 30 correct. These steps leave 493 tests on 143 tasks; the tests reject 264 of the 484 false positives of \S\ref{sec:rq1}. The other 81 tasks with false positives stay unhardened, most often because all of the task's passing patches other than the gold patch are false positives, so none of them can certify a test. The other half, the \emph{analysis half}, is held out for Appendix~\ref{app:rq4}.

\section{Verdict Quality}
\label{app:audit}

\paragraph{Human validation.} Two of the authors labelled 105 items independently, blind to the pipeline's verdicts and without AI assistance. The items were drawn at random within strata of audit outcome and difficulty, one per task. An FN item shows a task's statement and one marked F2P assertion, with the repository at the base commit. It asks whether the statement or the repository determines what the assertion checks, or whether this is left open (underdetermined). Of the 75 FN items, 45 are flagged assertions and 30 are controls. Each control is the assertion of a kept task that the base model's failing patches fail most often. An FP item shows a passing patch and asks whether it is correct. Both annotators judged 33 of the 45 flagged assertions left open: all 8 from never-solved tasks and 25 of the 37 from sometimes-solved ones (68\%). The two annotators judged 3 and 1 of the 30 controls left open and 10 and 12 of the 15 false positives incorrect. Each judged 1 of the 15 patches labelled correct to be incorrect. Cohen's $\kappa$ between them is 0.89 on FN items and 0.86 on FP items.

\paragraph{Agreement with PAIChecker.} Three of the patterns that PAIChecker's two annotators label~\citep{wang2026paichecker} describe tests or patches that go beyond the issue: unspecified literal, scope creep, and incomplete specification. Table~\ref{tab:quality} compares our verdicts with these labels on the 1{,}875 tasks that have both. On sometimes-solved tasks, the verdicts and the labels agree less: the audit gives an FN verdict to 46\% of the 123 tasks that the annotators mark and to 25\% of the others. PAIChecker labels a task as a whole, whereas we rate individual assertions against the statement and the repository.

\paragraph{The difficulty contrast under other definitions.} Table~\ref{tab:contrast} repeats the contrast of \S\ref{sec:rq1} under other definitions. Among the never-solved tasks of \Bset's pool, 79.2\% of those that stay unsolved in 16 further samples with a larger budget are FN-flawed, compared with 34.4\% of those solved at least once.

\begin{table}[h]
\centering
\caption{\textbf{The difficulty contrast under other definitions.} Share (\%) of flawed tasks among the never-solved and among the sometimes-solved tasks of each row.}
\label{tab:contrast}
\footnotesize
\begin{tabular}{@{}lcc@{}}
\toprule
 & Never solved & Sometimes solved \\
\midrule
FN-flawed, \Bset (\S\ref{sec:rq1}) & 74.0 & 19.5 \\
\quad counting repaired tasks as flawed & 83.1 & 28.0 \\
\quad counting only verdicts confirmed by both families & 67.5 & 12.9 \\
\quad tasks with one F2P test & 61.5 & 12.3 \\
\quad tasks with two or three F2P tests & 77.8 & 24.4 \\
\quad tasks with four or more F2P tests & 83.3 & 31.8 \\
FN-flawed, \Bset's pool (1{,}413 tasks) & 71.3 & 18.2 \\
Marked by PAIChecker's annotators, \Bset's pool (1{,}252 labelled tasks) & 61.0 & 20.6 \\
\bottomrule
\end{tabular}
\end{table}

\paragraph{Evidence behind FN verdicts.} Table~\ref{tab:tiers} splits the FN verdicts by the evidence behind them; ``other confirmed'' are verdicts reached in witness mode without two witnesses that fail a common F2P test. The 1{,}091 verdicts are seven fewer than the tasks that Fig.~\ref{fig:pipeline}b shows as removed or repaired after FN detection: the figure counts eight tasks that went through repair without an FN verdict, and it does not count one task with a verdict that was repaired by relaxing a test. In witness mode, execution shows that the tests are sensitive to the flagged detail, but only the family that wrote a witness judges whether it is a reasonable reading of the statement; only full escalation has both families judge the same executed patch. The executed witnesses differ from a passing fix in little else: 98.3\% pass every P2P test, and 92.1\% also pass every F2P test that contains no flagged assertion.

\begin{table}[h]
\centering
\caption{\textbf{Evidence behind FN verdicts (removed and repaired tasks).} Of the 213 verdicts that are confirmed by full escalation or contested, 203 come from the 374 tasks in full escalation, six from witness mode, and four from cleared tasks that were escalated afterwards (a random sample and every cleared never-solved task).}
\label{tab:tiers}
\footnotesize
\begin{tabular}{@{}lr@{}}
\toprule
Tier & Tasks \\
\midrule
Confirmed by full escalation (witnesses, cross-review, debate) & 85 \\
Witness mode: both families' witnesses fail a common F2P test & 810 \\
Other confirmed & 68 \\
Contested (one family) & 128 \\
\midrule
Total with an FN verdict & 1{,}091 \\
\bottomrule
\end{tabular}
\end{table}

\paragraph{Agreement between families.} The families agree on which tasks have a flagged assertion at $\kappa=$~0.80 (0.71 on \Bset's tasks). Claude's rating alone agrees with the final verdict on 91.0\% of the 2{,}001 tasks and GPT's on 94.8\%. Full escalation clears 171 of its 374 tasks. For false positives, Claude labels 705 and GPT 1{,}088 of the 1{,}854 passing patches incorrect ($\kappa=$~0.50). Both label 656 incorrect, and 484 of these have an executed counterexample. Of 79 cleared tasks escalated at random to witnesses and debate, one receives a confirmed FN verdict and a second a contested one. Neither is a sometimes-solved task.

\section{Training Sets}
\label{app:sets}

\paragraph{\Bset.} The pool holds the SWE-Gym tasks with a complete difficulty measurement (eight samples) under the original statements and tests, excluding the always-solved ones: 1{,}413 tasks. It is de-duplicated and already excludes pre-screen removals and the pull requests that PAIChecker labels defective. A statement-quality filter also drops statements that \mbox{GPT-5.6~Sol} rated underspecified (2 or more on SWE-bench Verified's 0--3 scale) or dependent on external resources. Within each difficulty level, tasks are ranked by a fixed hash of their IDs and taken in order: 77 never-solved tasks and 435 sometimes-solved ones, the latter allocated to \band{1}--\band{7} in proportion to the pool. \Bset keeps the original statements and tests.

\paragraph{\Aset.} \Aset is drawn from the 614 released tasks that are at \band{0}--\band{7} on the final statement and tests and whose audit had finished when \Aset was sampled. Tasks are taken in \Bset's rank order, with \Bset's quotas in four difficulty ranges (\band{0}, \band{1}--\band{2}, \band{3}--\band{5}, \band{6}--\band{7}). Where a range is short of tasks, the next-ranked tasks of the higher ranges fill it (15 tasks at \band{1}--\band{2}). All construction rules were fixed before \Aset was sampled, when no \Aset training run or held-out result existed. \Aset's difficulty is measured on final statements, \Bset's on the original ones.

\paragraph{\BF.} \BF applies a pass-rate filter to \Bset's pool: it has no never-solved task. Its \Nset tasks are allocated to \band{1}--\band{7} in proportion to the pool and taken in \Bset's rank order, so it contains \Bset's 435 sometimes-solved tasks and the 77 that follow them. It shares 336 tasks with \Aset, and 97 of its tasks are FN-flawed. We train on it with \Bset's seeds, in its own task order.

\begin{table}[h]
\centering
\caption{\textbf{The training sets.} Top: composition (release: trainable tasks, \band{0}--\band{7}). Bottom: why a task is in only one of \Aset and \Bset (the first reason that applies). Training reward: mean $\pm$ standard deviation over seeds. TV: total variation.}
\label{tab:sets}
\footnotesize
\begin{tabular}{@{}lcccc@{}}
\toprule
 & \Aset & \Bset & \BF & \SGV (trainable) \\
\midrule
Tasks (never / sometimes solved) & 77 / 435 & 77 / 435 & 0 / 512 & 138 / 570 \\
Repaired statements & 42 (8.2\%) & 0 & 0 & 20.9\% \\
Mean base pass rate & 49.8\% & 48.3\% & 56.8\% & \\
Median statement length (chars) / F2P tests & 2{,}353 / 1 & 2{,}040 / 2 & & \\
Repository mix, \Aset vs.\ \Bset (TV distance) & \multicolumn{2}{c}{5.3\%} & & \\
Training reward (\%) & 57.8\pmstd{0.7} & 52.1\pmstd{2.0} & 60.6\pmstd{1.8} & \\
\quad on the 314 tasks shared by \Aset and \Bset & 68.0\pmstd{1.7} & 66.0\pmstd{2.5} & & \\
Gradient-carrying groups & 65.9\% & 61.6\% & 70.8\% & \\
\midrule
\multicolumn{5}{@{}l}{\emph{\Bset only} (198): removed by the audit (156); audit unfinished when \Aset was sampled (30);} \\
\multicolumn{5}{@{}l}{\quad \band{8} after repair (11); other (1).} \\
\multicolumn{5}{@{}l}{\emph{\Aset only} (198): not in \Bset's pool (53); difficulty changed after repair (14);} \\
\multicolumn{5}{@{}l}{\quad beyond \Bset's quota in rank order (131).} \\
\bottomrule
\end{tabular}
\end{table}

\section{Training and Evaluation Settings}
\label{app:settings}

Table~\ref{tab:settings} lists the training and evaluation settings; the released configuration files give the remaining hyperparameters.

\begin{table}[h!]
\centering
\caption{\textbf{Training and evaluation settings.}}
\label{tab:settings}
\footnotesize
\begin{tabular}{@{}ll@{}}
\toprule
Base model & Qwen3.5-9B \\
Scaffold & SWE-agent \\
Episode budget & 80 tool calls, 128K context, temperature 1.0 \\
RL algorithm & GRPO with the Dr.~GRPO loss~\citep{liu2025drgrpo}; KL penalty to the base model, coefficient $10^{-3}$ \\
Optimization & Adam, learning rate $2\times10^{-6}$ (constant after 5 warm-up steps), weight decay 0.01, \\
 & \quad gradient-norm clip 1.0; 16 tasks $\times$ 8 samples per step, 80 steps \\
Seeds & seeds 1--3 for every set; \AFP seed $K$ uses the data order of \Aset seed $K$ \\
Checkpoint selection & highest dev-100 pass@1 (4 attempts) over steps 10--80, ties to the later step \\
dev-100 & 100 SWE-bench Verified tasks, fixed before any GRPO run, over-sampling tasks that the base model \\
 & \quad solves about half the time; pass@1 over 4 attempts, every 10 steps \\
\Vfour & the other 400 SWE-bench Verified tasks, 397 scored (two pass with an empty patch, one exceeds the \\
 & \quad test time limit); pass@1 over 4 attempts, training scaffold \\
SWE-bench Pro & public set of version~2~\citep{dan2026swebenchprov2}, 638 of 642 tasks scored (four gold patches fail in our \\
 & \quad harness); official SWE-agent tool-use scaffold; 150 calls, 256K context; 2 attempts \\
Statistics & means over three seeds $\pm$ sample standard deviation; gains over the base model paired by task \\
Compute & one node of 8 GPUs per run; 168--401 GPU-hours per run on the three matched sets (training only) \\
Training stack & Miles with Megatron-LM (full-parameter updates) and SGLang rollouts\anon{}{; AMD Instinct MI355X GPUs} \\
Audit cost & at least \$1.3 per audited task for LLM use (the GPT family's agent runs are not priced) \\
\bottomrule
\end{tabular}
\end{table}

\section{Additional Results for RQ2--RQ4}
\label{app:budget}\label{app:results}\label{app:rq4}

\paragraph{Label errors in the training signal.} Over the three GRPO runs on \Bset, 2{,}365 of 3{,}840 groups carry gradient: 461 of them (19.5\%) fall on sometimes-solved FN-flawed tasks (12.2\% counting only verdicts confirmed by both families) and 21 on never-solved ones. All rewards are zero in 20.7\% of \Bset's groups (\Aset: 13.9\%, \BF: 8.3\%), and 68.8\% of \Bset's all-zero groups fall on FN-flawed tasks.

\paragraph{Re-grading \Bset's zero-reward samples.} For every zero-reward training sample on \Bset's FN-flawed tasks, we mask the flagged assertions in the hidden tests, re-run the tests on the sample's patch, and count the samples that now pass (seeds 1 and 3; seed~2 has no test-level records). The subset used in \S\ref{sec:rq2} consists of the tasks on which every flagged assertion can be masked, the gold patch still passes, the empty patch still fails, and no masked check is repeated unmasked. On its 29 tasks, 268 of 864 samples pass (31.0\%); 28 of the 52 gradient-carrying groups and 35 of the 76 all-zero groups contain such a sample. On its 20 sometimes-solved tasks, 153 of 515 samples pass; 26 of the 49 gradient-carrying groups and 16 of the 36 all-zero groups contain one. Seed~3 stored no patch for 65 zero-reward samples of these tasks. They are not counted, and neither are the 11 groups that contain them. Over all 142 FN-flawed tasks, 19.5\% of the samples pass (706 of 3{,}613). This share is a lower bound, because 104 of the 221 flagged assertions cannot be masked mechanically.

\begin{table}[h]
\centering
\caption{\textbf{Per-run results.} pass@1 in \%; $\Delta$: paired difference from the base model (pp). Pro: two attempts. ckpt: selected step (dev-100 pass@1 at that step, the best of the run's eight checkpoints). The three \AFP runs happen to have the same dev-100 score. $\Delta$ is computed from unrounded values. The \AFP runs belong to RQ4.}
\label{tab:main}
\footnotesize
\setlength{\tabcolsep}{7pt}
\begin{tabular}{@{}lccccc@{}}
\toprule
Run & ckpt (dev-100) & \Vfour & \Vfour $\Delta$ & Pro & Pro $\Delta$ \\
\midrule
Base & -- (46.8\%) & 57.2\% & -- & 53.8\% & -- \\
\Aset s1 & 70 (57.5\%) & 63.8\% & $+$6.6 & 57.4\% & $+$3.6 \\
\Aset s2 & 60 (52.5\%) & 61.7\% & $+$4.5 & 54.8\% & $+$1.0 \\
\Aset s3 & 70 (54.5\%) & 62.3\% & $+$5.2 & 56.5\% & $+$2.7 \\
\Bset s1 & 20 (52.2\%) & 60.1\% & $+$3.0 & 51.6\% & $-$2.1 \\
\Bset s2 & 30 (55.8\%) & 62.3\% & $+$5.1 & 53.2\% & $-$0.5 \\
\Bset s3 & 20 (50.0\%) & 58.1\% & $+$0.9 & 55.7\% & $+$2.0 \\
\BF s1 & 80 (51.0\%) & 58.9\% & $+$1.8 & 48.0\% & $-$5.7 \\
\BF s2 & 40 (54.8\%) & 62.7\% & $+$5.5 & 53.6\% & $-$0.2 \\
\BF s3 & 50 (53.0\%) & 62.3\% & $+$5.1 & 52.6\% & $-$1.2 \\
\AFP s1 & 60 (53.5\%) & 63.0\% & $+$5.8 & 55.6\% & $+$1.8 \\
\AFP s2 & 40 (53.5\%) & 61.4\% & $+$4.2 & 54.9\% & $+$1.1 \\
\AFP s3 & 80 (53.5\%) & 63.3\% & $+$6.1 & 61.6\% & $+$7.8 \\
\bottomrule
\end{tabular}
\end{table}

\paragraph{Comparisons across seeds.} Averaged over the two benchmarks, the three \Aset runs gain $+5.1$, $+2.8$, and $+4.0$\pp over the base model and the three \Bset runs $+0.4$, $+2.3$, and $+1.5$\pp, so every \Aset run gains more than every \Bset run. On each benchmark alone, the runs of the two sets overlap, but the difference between the seed means of \Aset and \Bset remains at least 1.4\pp when any single run is left out. \Aset outperforms \BF by 1.3\pp on \Vfour, by 4.8\pp on Pro, and by 3.1\pp on their mean. \Aset's lead on the mean remains at least 1.6\pp when any single run is left out. Whether \BF gains more than \Bset depends on one run. Without \BF's seed~1, which is 5.7\pp below the base model on Pro, \BF's average gain is $+2.3$\pp instead of $+0.9$\pp, above \Bset's $+1.4$\pp. On dev-100 at the last checkpoint (step~80), the \Aset runs score 54.8, 50.2, and 54.0\% and the \Bset runs 49.5, 45.2, and 49.2\%.

\paragraph{False positives during RL.} In a random sample of 300 passing patches from the analysis half (Appendix~\ref{app:fp}; \Aset seeds 1 and~2), 87 (29.0\%) are false positives, 97 are labelled correct by both families, and 116 are undetermined. The new tests reject 25.3\% of the sampled false positives; another 51.7\% fall on tasks without new tests and therefore cannot be caught. On the hardened tasks, the tests reject 18.9\% of the analysis half's passing patches, and both families judge 96.1\% of the rejected patches incorrect.

In each 20-step bin of Fig.~\ref{fig:fp}a, the re-graded \Aset reward is \Aset's reward times the analysis half's acceptance rate under the new tests, with each distinct patch weighted by the number of training episodes that produced it. Re-grading lowers \Aset's reward on the hardened tasks by 13.2 and 14.5\pp in seeds 1 and~2. The estimate assumes that the analysis half is representative of all of \Aset's passing patches. With this estimator, the \AFP policy passes the new tests 0.4 and 1.8\pp more often than the \Aset policy would.

\section{Case Studies}
\label{app:cases}

\paragraph{A wording test inside GRPO.} \emph{pandas-dev/pandas-53409} (confirmed by both families; removed): the issue asks to deprecate literal JSON input to \texttt{read\_json} in favour of \texttt{StringIO} and gives no warning text. The F2P test requires a \texttt{FutureWarning} matching ``Passing literal json to `read\_json' is deprecated and will be removed in a future version. To read from a literal string, wrap it in a `StringIO' object.'' Both families' witnesses word the warning differently and fail at this match; the base model passes in two of eight samples. In the task's six GRPO visits in the two \Bset runs with test-level records, 8 samples are rewarded and 15 others fail only this test; 11 of the 15 pass once the match on the message is removed.

\paragraph{A never-solved flawed task inside GRPO.} \emph{getmoto/moto-5137} (confirmed by both families; removed): the issue asks for pagination of \texttt{list\_accounts\_for\_parent}. One F2P test also requires pagination of \texttt{list\_organizational\_units\_for\_parent}, which the issue never mentions; both families' witnesses paginate only the first API and fail there, and the base model never solves the task. In its five GRPO visits in the two \Bset runs with test-level records, 30 of the 40 samples fail only the test of the second API. The single rewarded sample is the only one whose patch touches the second API: four groups carry no gradient, and the fifth rewards doing more than the issue asks.

\paragraph{A repair.} \emph{getmoto/moto-4846} (confirmed by both families; a repaired task of \Aset): the issue reports that removing a permission that does not exist raises no error, whereas AWS raises \texttt{ResourceNotFoundException}. The F2P test also requires the error message to equal ``No policy is associated with the given resource.'', which the issue does not give; both families' witnesses raise the exception with another message and fail. The repair appends one requirement, which states this message. The base model solves the original task in none of eight samples and the repaired task in seven. In \emph{iterative/dvc-4005} the missing requirement is an entry point. The issue concerns metrics and plots, but the F2P test calls \texttt{DvcTree.open} directly, so the repair has to name this entry point, which also tells the agent where the fix belongs. The pass count again rises from 0 to 7 of 8.

\paragraph{A contested verdict.} \emph{Project-MONAI/MONAI-2454} (contested; removed after a repair attempt): the issue reports that \texttt{ToTensor()} adds a dimension to ``a numpy array with 0 dims'' and gives the root cause, \texttt{np.ascontiguousarray}, with the example \texttt{np.ascontiguousarray(2)}. The F2P test also calls \texttt{ToTensor()} on the Python integer 5. A patch that special-cases zero-dimensional \texttt{np.ndarray} inputs fixes the array case but fails the test on the integer. One family holds that the issue reports arrays only, so the integer goes beyond it; the other holds that the issue names the root cause and shows it on an integer, so a fix that leaves integers wrong does not resolve the issue.

\paragraph{False positives that the new tests miss.} \emph{pydantic/pydantic-8650} (hardened in \AFP): the issue asks that \texttt{NameEmail} validate from JSON. Five of the seven distinct passing patches are false positives. Of the five tests built from their counterexamples, three are discarded because they reject a further passing patch that only one family labels incorrect, and one because it rejects a training patch on which the families disagree. The test that remains rejects one of the five false positives. In the task of Fig.~\ref{fig:teaser}b, the new tests reject all five false positives among six passing patches.

\section{Terms}
\label{app:terms}

\begin{table}[h]
\centering
\caption{\textbf{Terms.}}
\label{tab:terms}
\footnotesize
\begin{tabular}{@{}l>{\raggedright\arraybackslash}p{0.58\textwidth}l@{}}
\toprule
Term & Meaning & Defined in \\
\midrule
F2P / P2P tests & tests that must start passing / keep passing & \S\ref{sec:defs} \\
Pass count $k$ & passes in 8 base-model samples: never ($0$), sometimes ($1$--$7$), always ($8$) solved & \S\ref{sec:defs} \\
FN / FP & correct patch rejected / wrong patch accepted & \S\ref{sec:defs} \\
Family & Claude Opus~5.5 with Claude~Code, or GPT-6.1~Sol with Codex~CLI. \mbox{GPT-5.6~Sol} also counts as GPT & \S\ref{sec:pipeline} \\
Flagged assertion & an F2P assertion that a family judges may reject a correct patch & \S\ref{sec:fn} \\
Witness & a patch that follows the statement but differs from the gold patch in the flagged detail & \S\ref{sec:fn} \\
FN verdict & given to a task when a flagged assertion rejects a witness that the authoring family judges reasonable; confirmed (both families) or contested (one) & \S\ref{sec:fn} \\
Repaired / FN-flawed & a task with an FN verdict that a statement repair fixes / that is removed & \S\ref{sec:defs} \\
Gradient-carrying group & a GRPO group whose rewards are not all equal & \S\ref{sec:defs} \\
\Bset{} / \Aset{} / \AFP{} / \BF & original tasks / audited tasks / audited tasks with the new tests / original tasks, with never-solved tasks replaced by sometimes-solved ones & \S\ref{sec:setup} \\
\Vfour{} / dev-100 & 397 held-out SWE-bench Verified tasks / 100 other SWE-bench Verified tasks, used for checkpoint selection & \S\ref{sec:setup} \\
\bottomrule
\end{tabular}
\end{table}